\documentclass[10pt, a4paper]{article}

\usepackage{lrec-coling2024} 

\usepackage{graphicx}
\usepackage{subcaption}
\usepackage{enumitem}
\usepackage{times}
\usepackage{latexsym}
\usepackage{hhline}

\usepackage[T1]{fontenc}

\usepackage[utf8]{inputenc}

\usepackage{microtype}

\usepackage{inconsolata}
\usepackage{graphicx}
\usepackage{multirow}
\usepackage{amsfonts} 
\usepackage{mathrsfs}

\title{\textbf{Beyond the Known: Investigating LLMs Performance on Out-of-Domain Intent Detection}}

\name{Pei Wang$^{1*}$, Keqing He$^{2*}$, Yejie Wang$^{1*}$, Xiaoshuai Song$^{1}$, Yutao Mou$^{1}$, \\ \thanks{\ * The first three authors contribute equally. Weiran Xu is the corresponding author.}
\textbf{\large{Jingang Wang$^{2}$, Yunsen Xian$^{2}$, Xunliang Cai$^{2}$, Weiran Xu$^{1*}$}}}

\address{
         \{wangpei, wangyejie, songxiaoshuai,myt,xuweiran\}@bupt.edu.cn\\
         \{hekeqing,wangjingang,xianyunsen,caixunliang\}@meituan.com}

\abstract{
Out-of-domain (OOD) intent detection aims to examine whether the user's query falls outside the predefined domain of the system, which is crucial for the proper functioning of task-oriented dialogue (TOD) systems. Previous methods address it by fine-tuning discriminative models. Recently, some studies have been exploring the application of large language models (LLMs) represented by ChatGPT to various downstream tasks, but it is still unclear for their ability on OOD detection task.This paper conducts a comprehensive evaluation of LLMs under various experimental settings, and then outline the strengths and weaknesses of LLMs. We find that LLMs exhibit strong zero-shot and few-shot capabilities, but is still at a disadvantage compared to models fine-tuned with full resource. More deeply, through a series of additional analysis experiments, we discuss and summarize the challenges faced by LLMs and provide guidance for future work including injecting domain knowledge, strengthening knowledge transfer from IND(In-domain) to OOD, and understanding long instructions.
 \\ \newline  }

\begin{document}

\maketitleabstract

\section{Introduction}

Traditional TOD systems are based on the closed-set hypothesis \citep{chen2019bert,yang2021generalized,zeng-etal-2022-semi} and can only handle queries within a limited scope of in-domain(IND) intents. However, users may input queries with out-of-domain(OOD) intents in the real open world, which poses new challenges for TOD systems.  As shown in Figure \ref{intro}, OOD intent detection task aims to determine whether the intent of user queries exceeds the predefined intents, making it an essential component of TOD systems. \cite{Tulshan2018SurveyOV, Lin2019DeepUI, zeng-etal-2021-modeling, Wu2022RevisitOF, Wu2022DistributionCF, Mou2022UniNLAR,mou-etal-2023-decoupling,song-etal-2023-continual}.

\begin{figure}[t]
  \centering
  \includegraphics[width=0.5\textwidth]{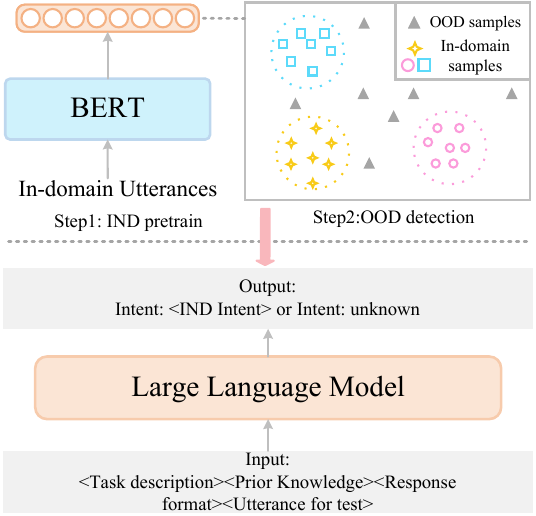}
  \caption{Comparison of the OOD detection method between previous method (Upper part) and LLM-based method (Lower part). Previous method trains a feature extractor using IND samples in the first stage, and estimates the confidence score of the sample using the designed scoring function and features; Our end-to-end OOD detection based on LLM adds task descriptions to prompts, and LLM directly outputs detection results.}
  \label{intro:method}
\end{figure}

Previous work on OOD detection rely on fine-tuning pre-training language model (PLM), extracting the output representation of  PLMs’ final layer as the intent feature, and employing scoring functions based on density, distance, or energy to detect OOD samples, as shown in Figure \ref{intro:method} \citep{zeng-etal-2021-modeling, zhou-etal-2022-knn, Mou2022UniNLAR,cho2023probing, wang-etal-2023-app}. Recently, the emergence of large language models(LLMs) like ChatGPT\footnote{https://openai.com/blog/ChatGPT} has injected new vitality into natural language process(NLP) tasks. Their superior zero-shot learning capability enables a new paradigm of NLP research and applications by prompting LLMs without finetuning \cite{Ouyang2022TrainingLM,Touvron2023LLaMAOA,jiao2023chatgpt,wei2023zeroshot,yang2023exploring}. Given the LLMs' training on broad text corpora and their impressive generalization skills, it's worth considering the benefits and potential challenges they may face in open-scenario intent identification. Specifically, we have raised the following questions:

\begin{figure}[h]
  \centering
  \includegraphics[width=0.5\textwidth]{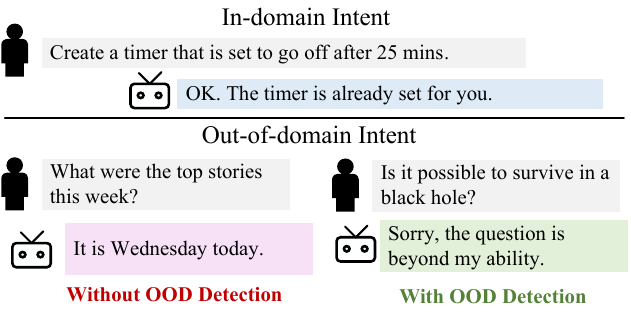}
  \caption{Explanation of the role of OOD intent detection in the TOD system. When the system encounters an intent that is beyond its supported intents, it can detect and friendly prompt the user.}
  \label{intro}
\end{figure}

\begin{enumerate}
[itemsep=4pt,topsep=0pt,parsep=0pt,leftmargin=9pt]
\item What are the potential positive and negative effects of large language models on the Out-of-Domain (OOD) detection task?

\item What are the strengths and weaknesses of large language models, compared with traditional fine-tuned models?

\item Why do large language models exhibit certain strengths and weaknesses?

\item How can we potentially address and improve these weaknesses?
\end{enumerate}

In this work, we introduce two LLM-based OOD framework, ZSD-LLM and FSD-LLM which based on different IND prior to instruct LLM to conduct intent detection (Section \ref{sec:method}). Then we conduct comparative experiments between ChatGPT and discriminative methods (Section \ref{sec:experiment}). In order to further explore the underlying reasons behind the experiments, we conduct a series of analytical experiments including IND intent number effect, different data split, comparison of different LLMs and different prompts effect (Section\ref{sec:qualitative}). Finally, we summarize the strengths and weeknesses of ChatGPT in OOD detection tasks and future improvement directions (Section \ref{sec:discussion}). To the best of our knowledge, we are the first to comprehensively evaluate the performance of LLMs on OOD intent detection. 

The key findings of this paper can be summarized as follows:


\textbf{What ChatGPT does well:}
\begin{itemize}
[itemsep=4pt,topsep=0pt,parsep=0pt,leftmargin=9pt]
\item ChatGPT can achieve good zero-shot performance without providing any IND intent priors, demonstrating his powerful NLU capabilities.
\item When the number of IND intents is small, ChatGPT can achieve better accuracy  in few-shot settings than discriminative models.
\item ChatGPT can not only perform OOD detection but also output the intent of the OOD samples, which is something that current methods based on discriminative models cannot achieve.
\end{itemize}

\textbf{What ChatGPT does not do well:}
\begin{itemize}
[itemsep=4pt,topsep=0pt,parsep=0pt,leftmargin=9pt]
\item ChatGPT performs significantly worse than baselines with a large number of IND intents. It's manifested by an increase in misclassifications among IND intents and a substantial number of OOD samples being detected as IND when there is a higher number of IND intents.
\item In rare instances, ChatGPT does not output according to our designed instructions. Particularly when the increase in intents leads to longer instructions, ChatGPT may overlook key information in the prompts, resulting in task failure.
\item Compared to discriminative models, the performance of ChatGPT is affected to some extent by the number of intents. This is primarily manifested when the number of IND intents increases, resulting in a significant decline in its performance.
\item ChatGPT struggles with fine-grained semantic distinctions which indicating the comprehension of ChatGPT in fine-grained intent labels is insufficient, exhibiting misalignment with human-level understanding.
\item It's challenging for ChatGPT to acquire knowledge from IND demonstrations that could assist with OOD tasks. It might even perceive IND demonstrations as noisy, which could potentially harm the performance of OOD tasks.
\end{itemize}

We further summarize future LLM improvement directions which includes the following aspects: 1) injecting domain knowledge 2) strengthening knowledge transfer from IND to OOD, and 3) understanding long instructions.\footnote{Our codes can be found on \textit{https://github.com/Yupei-Wang/ood\_llm\_eval}}

\section{Related Work}
\label{sec:append-how-prod}
\subsection{LLM}
LLM has become a popular paradigm for research and applications in natural language processing tasks. ChatGPT is a generative foundational model belonging to the GPT-3.5 series in the OpenAI GPT family, which includes its predecessors, GPT, GPT-2, and GPT-3. Recently, there has been an increasing interest in utilizing LLMs for various natural language processing (NLP) tasks. Several studies have been conducted to systematically investigate the performance of ChatGPT on different downstream tasks, including machine translation \cite{jiao2023chatgpt},information extraction \cite{wei2023zeroshot}, summarization\cite{yang2023exploring} and clustering \cite{song-etal-2023-large}. However, it is unclear about the performance of ChatGPT in OOD detection. 

\begin{figure*}[t]
  \centering
  \vspace{-10pt}
  \includegraphics[width=1.0\textwidth]{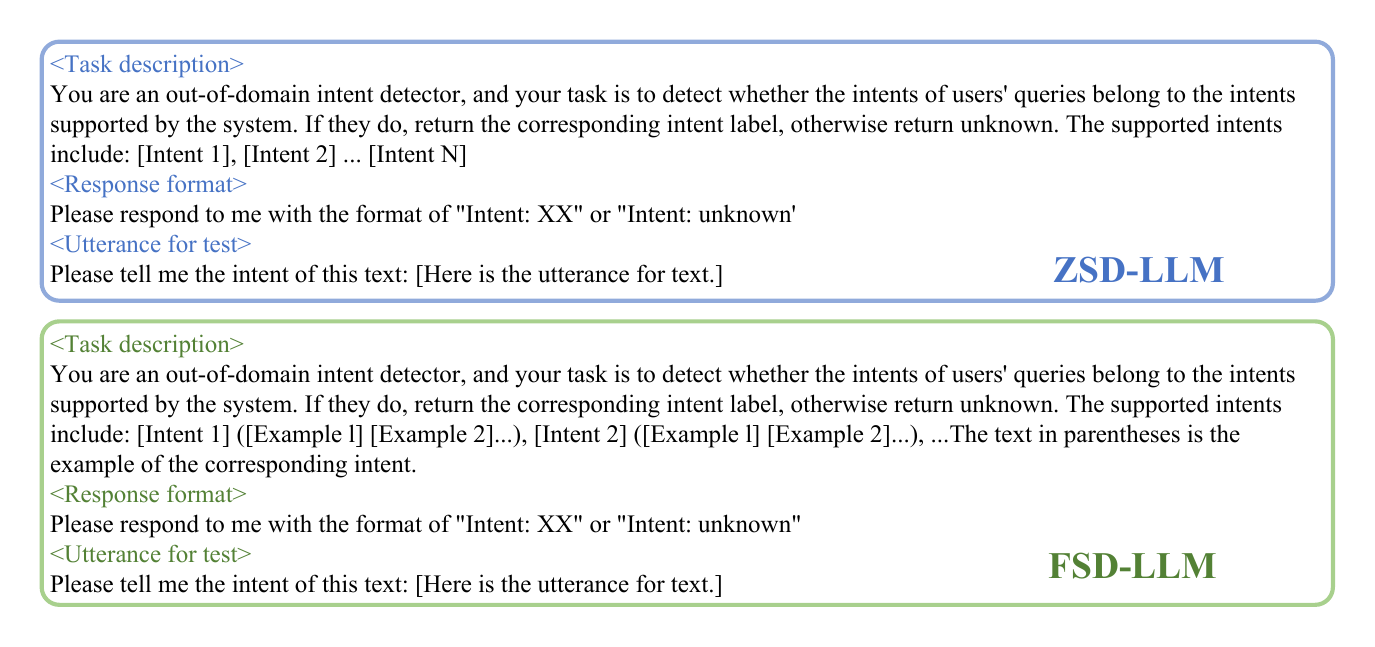}
  \caption{The demonstration of the two prompts we use to assist ChatGPT in performing OOD intent detection. FSD-OOD incorporates examples of intentions in the prompt as prior knowledge.}
  \label{method_prompts}
\end{figure*}

\subsection{OOD Detection}

Previous OOD detection methods can be divided into two categories: supervised OOD detection \cite{Fei2016BreakingTC,Kim2018JointLO,Larson2019AnED,Zheng2020OutofDomainDF}  and unsupervised OOD detection\cite{Shu2017DOCDO,Lee2018ASU,Ren2019LikelihoodRF,Lin2019DeepUI,xu-etal-2020-deep,zeng-etal-2021-modeling, Mou2022UniNLAR}. The former indicates that there are some extensive labeled OOD samples in the training data. Classic supervised OOD algorithms consider the OOD detection problem as an N+1 classification problem \cite{Fei2016BreakingTC,Larson2019AnED}. Unsupervised methods generally perform in two stages: learning intent representation and estimating confidence scores \cite{Mou2022UniNLAR}. Most previous research has focused on fine-tuning small-scale pre-trained language models (PLMs), such as BERT, to learn intent features from the training data \cite{wang-etal-2023-app}. However, with the recent advancements in LLM, there is increasing interest in exploring their potential for OOD intent detection. Compared to small-scale PLMs, LLMs have greater capacity for learning and generalization from data, making them promising candidates for OOD intent detection.

\section{Methodology}
\label{sec:method}
\subsection{Problem Formulation}
Given predefined set of intents, denoted as $\mathcal S=\left\{l_{1}, l_{2}, \ldots, l_{N}\right\}$ 
,it contains N intents supported by the system. The input is the user's natural language query $q = \{t_1, t_2, \dots, t_n\}$, where $t_{i}$ represents $i$th token in the query. The output is an intent label $ l_{pre} $ that belongs to the set $ \mathcal {S} \cup \left\{ OOD \right\}$.

\subsection{Prompt Engineer}

We evaluate the OOD intent detection capability of ChatGPT in an end-to-end manner. We heuristically propose two prompts based on different IND prior:

\textbf{Zero-shot Detection (ZSD-LLM)}: This method only provides the IND intent set in the prompt as prior knowledge without supplying any IND samples. It can be utilized in scenarios where user privacy protection is required. The prompt template is: <Task description><Prior: $\mathcal {S}$><Response format><Utterance for test>.

\textbf{Few-shot Detection (FSD-LLM)}: This method provides several samples for each intent in the prompt, allowing ChatGPT to extract useful knowledge from these samples and apply it to distinguish between IND and OOD intents. The prompt template is: <Task description><Prior: $D = \{(q_1, l_1), (q_2,l_2) \dots, (q_n,l_n)\}$><Response format><Utterance for test>.

 We show this two methods in Figure \ref{method_prompts}. About the exploration of different prompts, we discuss it in Section \ref{sec:diff_prompt}.

\begin{table*}[t]
\resizebox{1\textwidth}{!}{%
\begin{tabular}{l|llllll|llllll|llllll}
\hline
\multicolumn{1}{c|}{\multirow{2}{*}{Model}}                                       & \multicolumn{6}{c||}{Split = 25\%}                                                                                                                                                                        & \multicolumn{6}{c||}{Split = 50\%}                                                                                                                                                                         & \multicolumn{6}{c}{Split = 75\%}                                                                                                                                                                        \\ \cline{2-19}  &
 \multicolumn{2}{c|}{ALL}                                    & \multicolumn{2}{c|}{IND}                                   & \multicolumn{2}{c||}{OOD}                                    & \multicolumn{2}{c|}{ALL}                                    & \multicolumn{2}{c|}{IND}                                    & \multicolumn{2}{c||}{OOD}                                    & \multicolumn{2}{c|}{ALL}                                    & \multicolumn{2}{c|}{IND}                                    & \multicolumn{2}{c}{OOD}                                    \\ \cline{2-19} 
\multicolumn{1}{c|}{}                                & ACC & \multicolumn{1}{c|}{F1} & ACC & \multicolumn{1}{c|}{F1} & Recall & \multicolumn{1}{c||}{F1} & ACC & \multicolumn{1}{c|}{F1} & ACC & \multicolumn{1}{c|}{F1} & Recall & \multicolumn{1}{c||}{F1} & ACC & \multicolumn{1}{c|}{F1} & ACC & \multicolumn{1}{c|}{F1} & Recall & \multicolumn{1}{c}{F1} \\ \hline
SCL                                          & 74.36        & \multicolumn{1}{l|}{61.06}       & 71.57        & \multicolumn{1}{l|}{60.18}       & 76.46       & \multicolumn{1}{l||}{77.76}                           & 75.45        & \multicolumn{1}{l|}{69.94}       & 80.43        & \multicolumn{1}{l|}{69.98}       & 67.44        & \multicolumn{1}{l||}{68.25}                            & 79.43             & \multicolumn{1}{l|}{ 84.54}            & 82.35             & \multicolumn{1}{l|}{84.86}            & 71.06             &        66.34                            \\ \hline
KNN-CL                                       & 88.21        & \multicolumn{1}{l|}{77.65}       & 78.16        & \multicolumn{1}{l|}{77.94}       &91.77        &\multicolumn{1}{l||}{92.36}                            & \textbf{81.98}        & \multicolumn{1}{l|}{\textbf{83.67}}       & \textbf{85.13}        & \multicolumn{1}{l|}{\textbf{83.72}}       & \textbf{85.25}        & \multicolumn{1}{l||}{\textbf{81.96}}                            & 81.69        & \multicolumn{1}{l|}{70.21}       & \textbf{86.30}         & \multicolumn{1}{l|}{86.03}       & 85.13        & 71.88                            \\ \hline
UniNL                                        & \textbf{89.41}        & \multicolumn{1}{l|}{\textbf{80.04}}       & 78.59        & \multicolumn{1}{l|}{\textbf{79.36}}       & \textbf{92.96}        & \multicolumn{1}{l||}{\textbf{93.02}}                            & 81.42        & \multicolumn{1}{l|}{82.66}       & 84.68        & \multicolumn{1}{l|}{82.70}        & 78.24        & \multicolumn{1}{l||}{81.18}                            & \textbf{82.78}        & \multicolumn{1}{l|}{\textbf{86.36}}       & 81.78        & \multicolumn{1}{l|}{\textbf{86.59}}       & \textbf{85.66}        & \textbf{73.34}                            \\ \hline \hline

ChatGPT                                      & 47.5         & \multicolumn{1}{l|}{42.68}       & 73.16        & \multicolumn{1}{l|}{42.02}       & 39.09        & \multicolumn{1}{l||}{55.17}                            & 46.46        & \multicolumn{1}{l|}{54.91}       & 71.32        & \multicolumn{1}{l|}{55.47}       & 22.24        & \multicolumn{1}{l||}{33.77}                            & 50.58        & \multicolumn{1}{l|}{56.97}       & 62.85        & \multicolumn{1}{l|}{57.58}       & 15.62        & 22.03                            \\ \hline
\end{tabular}
}
\caption{The performance comparison between ChatGPT and  baselines of Banking. We select 25\%, 50\%, and 75\% of all intents as IND intents. Three average values are taken for each experiment.}
\label{result_banking}
\end{table*}

\begin{table*}[t]
\resizebox{1\textwidth}{!}{%
\begin{tabular}{l|llllll|llllll|llllll}
\hline
\multicolumn{1}{c|}{\multirow{2}{*}{Model}}                                             & \multicolumn{6}{c||}{Split = 25\%}                                                                                                                                                                        & \multicolumn{6}{c||}{Split = 50\%}                                                                                                                                                                        & \multicolumn{6}{c}{Split = 75\%}                                                                                                                                                                         \\ \cline{2-19}
 & \multicolumn{2}{c|}{ALL}                                   & \multicolumn{2}{c|}{IND}                                    & \multicolumn{2}{c||}{OOD}                                    & \multicolumn{2}{c|}{ALL}                                    & \multicolumn{2}{c|}{IND}                                    & \multicolumn{2}{c||}{OOD}                                    & \multicolumn{2}{c|}{ALL}                                    & \multicolumn{2}{c|}{IND}                                    & \multicolumn{2}{c}{OOD}                                    \\ \cline{2-19} 
\multicolumn{1}{c|}{}                                & ACC & \multicolumn{1}{c|}{F1} & ACC & \multicolumn{1}{c|}{F1} & Recall & \multicolumn{1}{c||}{F1} & ACC & \multicolumn{1}{c|}{F1} & ACC & \multicolumn{1}{c|}{F1} & Recall & \multicolumn{1}{c||}{F1} & ACC & \multicolumn{1}{c|}{F1} & ACC & \multicolumn{1}{c|}{F1} & Recall & \multicolumn{1}{c}{F1} \\ \hline
SCL                                         & 87.64       & \multicolumn{1}{l|}{88.32}       & 91.44        & \multicolumn{1}{l|}{89.08}       & 74.11        & \multicolumn{1}{l||}{77.76}                            & 85.82        & \multicolumn{1}{l|}{83.64}       & 84.89       & \multicolumn{1}{l|}{83.58}       & 86.46      & \multicolumn{1}{l||}{87.94}                            & 89.18       & \multicolumn{1}{l|}{90.55}       & 88.75       & \multicolumn{1}{l|}{90.58}       & 89.86        & \textbf{87.13}                            \\ \hline
KNN-CL                                      & \textbf{92.04}        & \multicolumn{1}{l|}{83.31}       & 84.86        & \multicolumn{1}{l|}{82.99}       & \textbf{93.85}        & \multicolumn{1}{l||}{\textbf{94.97}}                            & 90.33        & \multicolumn{1}{l|}{88.52}       & 88.53        & \multicolumn{1}{l|}{88.47}       & \textbf{91.57}        & \multicolumn{1}{l||}{\textbf{91.95}}                            & 89.18        & \multicolumn{1}{l|}{92.03}          & 88.49        & \multicolumn{1}{l|}{92.10}        & \textbf{92.30}         & 77.66                            \\ \hline
UniNL                                        & 87.8         & \multicolumn{1}{l|}{\textbf{89.79}}       & \textbf{97.27}        & \multicolumn{1}{l|}{\textbf{89.89}}       & 77.3         & \multicolumn{1}{l||}{86.14}                            & \textbf{90.95}       & \multicolumn{1}{l|}{\textbf{93.15}}       & \textbf{95.79}        & \multicolumn{1}{l|}{93.25}  &    80.03      & \multicolumn{1}{l||}{85.75}                            & \textbf{91.77}        & \multicolumn{1}{l|}{\textbf{94.01}}       & \textbf{93.84}        & \multicolumn{1}{l|}{\textbf{94.09}}       & 83.95        & 82.74                            \\ \hline \hline

ChatGPT                                      & 63.86        & \multicolumn{1}{l|}{58.86}       & 81.26        & \multicolumn{1}{l|}{58.51}       & 58.15        & \multicolumn{1}{l||}{71.82}                            & 59.84       & \multicolumn{1}{l|}{69.9}        &82.4         & \multicolumn{1}{l|}{70.14}       & 37.29        & \multicolumn{1}{l||}{51.43}                            & 64.24        & \multicolumn{1}{l|}{70.18}       & 74.79        & \multicolumn{1}{l|}{70.44}       & 33.16        & 41.71                           \\ \hline
\end{tabular}
}
\caption{The performance comparison between ChatGPT and  baselines of CLINC. We select 25\%, 50\%, and 75\% of all intents as IND intents. Three average values are taken for each experiment.}
\label{result_clinc}
\end{table*}

\section{Experiment}
\label{sec:experiment}
\subsection{Setup}
\subsubsection{Dataset \& Metric}
\textbf{Dataset} We conduct experiments on two widely used benchmark, CLINC \citep{Larson2019AnED} and Banking \citep{casanueva2020efficient}. CLINC consists of 150 intents distributed across 10 domains, with each domain containing 15 intents. Banking contains intents from a single domain, totaling 77 intents. Consistent with previous research, we conduct OOD detection under three settings: 25\%, 50\%, and 75\%. Here, 25\% refers to selecting 25\% of the intents as IND, with the remaining intents considered as OOD. We show the detailed statistics of the datasets in Table \ref{dataset}. 

\textbf{Metric} We employ six commonly used OOD detection metrics to evaluate the performance, including IND metrics: accuracy and macro-F1, OOD metrics: recall and macro-F1, as well as overall accuracy and macro-F1.

\begin{table}
\resizebox{0.5\textwidth}{!}{%
\begin{tabular}{l|c|c}
\hline
Statistic                    & \multicolumn{1}{l|}{Banking} & \multicolumn{1}{l}{CLINC} \\ \hline
Avg utterance length         & 9                            & 12                         \\ \hline
Intent                       & 150                          & 77                         \\ \hline
Training set size            & 15000                        & 9003                       \\ \hline
Training sample per class    & 100                          & -                          \\ \hline
Development set size         & 3000                         & 1000                       \\ \hline
Development sample per class & 20                           & -                          \\ \hline
Testing set size             & 5500                         & 3080                       \\ \hline
Testing sample per class     & 30                           & -                          \\ \hline
\end{tabular}
}
\vspace{0cm}
\caption{Statistics of datasets.}
\label{dataset}
\end{table}

\subsubsection{Baselines}
We compare ChatGPT with the following three state-of-the-art discriminative two-stage methods:

\textbf{SCL} \cite{zeng-etal-2021-modeling} It proposes a supervised contrastive learning objective to minimize intra-class variance by pulling together in-domain intents belonging to the same class and maximize inter-class variance by pushing apart samples from different classes.

\textbf{KNN-CL} \cite{zhou-etal-2022-knn} It proposes a KNN-based contrastive loss for IND pre-training. KNN-CL selects k-nearest neighbors from samples of the same class as positives and uses samples of the different classes as negatives. 

\textbf{UniNL} \cite{Mou2022UniNLAR} It proposes a unified Neighborhood Learning to align representation learning with the scoring function to improve OOD detection performance. KNCL objective is employed for IND pre-training and a KNN-based score function is used for OOD detection. 

\subsection{ZSD-LLM Results}
\label{sec:main_result}

Our results are shown in Table \ref{result_banking} and \ref{result_clinc}. The results show that ZSD-LLM performs worse than the best baselines on all metrics. we analyze the results from three aspects:

(1) \textbf{The performance of IND intent recognition.} There is a certain gap between ChatGPT and strong baselines (UniNL, KNN-CL). Taking Banking-50\% as an example, in terms of IND indicators, ChatGPT's performance is 13.36\% (IND-ACC) and 27.23\% (IND-F1) lower than UniNL. However, the gap between ChatGPT and SCL is slightly smaller, and it even surpasses SCL in some settings, such as Banking 25\% (71.57 -> 73.16). This demonstrates ChatGPT's strong zero-shot capability.

(2) \textbf{The performance of OOD sample detection.} Compared to IND classification, the performance gap between ChatGPT and baselines is larger on OOD metrics. Specifically, ChatGPT's OOD-Recall is reduced by 56\%, and OOD-F1 is reduced by 47.41\% compared with UniNL for banking-50\%. It's generally observed across three IND intent splits in the two datasets. We speculate that the reason for the lower OOD metrics is that a large number of OOD samples are misclassified as IND intents by ChatGPT. Such results indicate that this kind of zero-shot prompting is not enough to provide ChatGPT with sufficient prior knowledge to complete the OOD detection task.

(3) \textbf{Comparison between datasets.} The performance comparison of ChatGPT between the two datasets shows the same trend as the baselines. The detection ability of the multi-domain dataset (CLINC) is better than that of the single-domain dataset (Banking). On the simpler dataset CLINC, the gap between ChatGTP and UniNL is smaller than that on Banking. (ALL-ACC: 41.91 -> 23.84 for 25\%, 34.96 -> 31.11 for 50\% ,32.20 -> 27.53 for 75\%). This indicates that the high granularity of intent division is the reason for the poor performance of ChatGPT.

(3) \textbf{Task fail.} In addition to the above results, we use the original OOD data from CLINC for OOD detection. This results in a total of 150 intents, which can be used to test ChatGPT's ability to perform large-scale system OOD detection. Experimental results show that ChatGPT predicts new intents in approximately 8.49\% of the test samples, neither returning an IND intent nor 'unknown', leading to task failure. This reflects the instability of LLM in performing OOD detection.

\begin{figure*}[t]
  \centering
  \begin{minipage}[t]{0.245\textwidth}
    \centering
    \includegraphics[width=\linewidth]{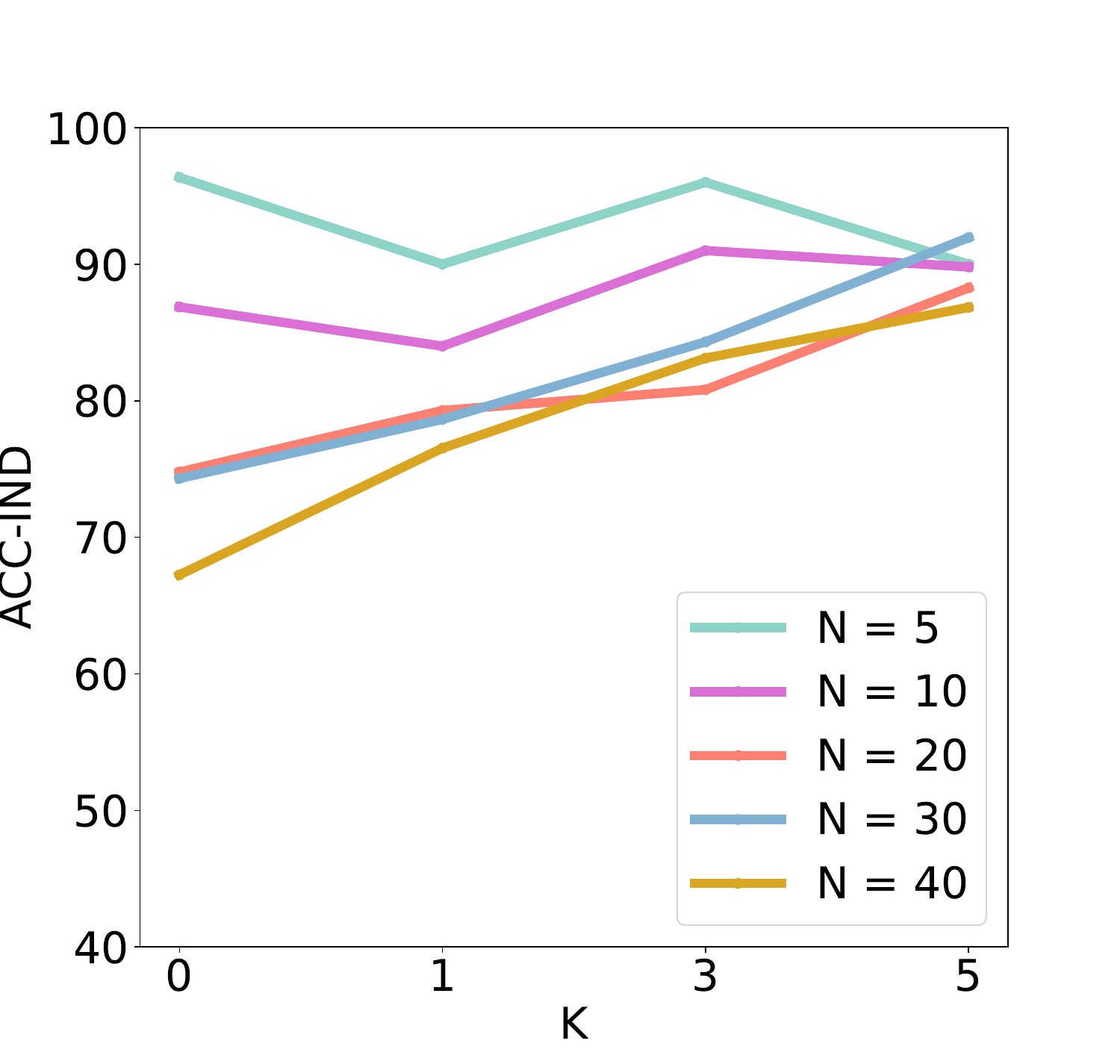}
   \subcaption[]{}
    \label{fig:sub1}
  \end{minipage}%
  \begin{minipage}[t]{0.245\textwidth}
    \centering
    \includegraphics[width=\linewidth]{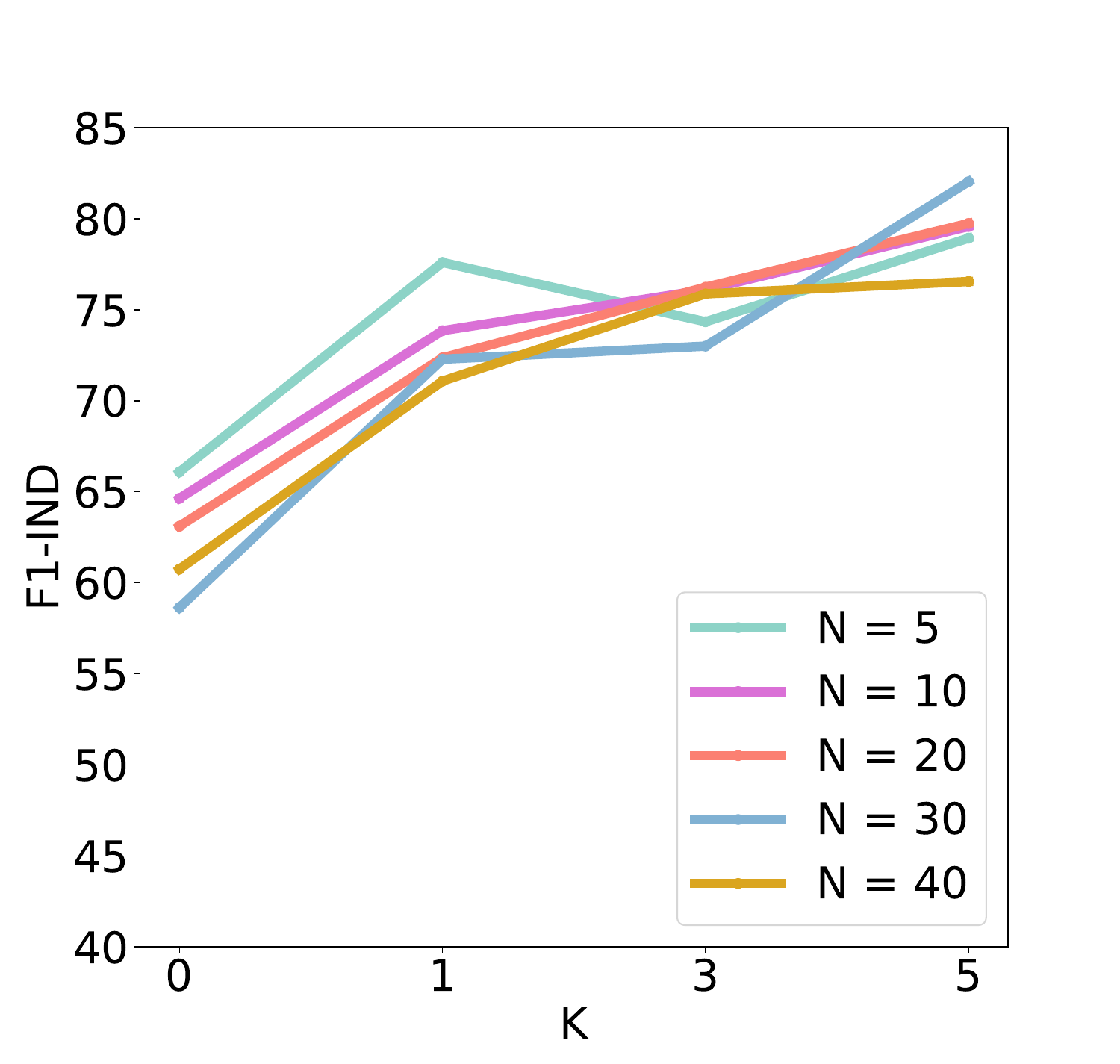}
    \subcaption[]{}
    \label{fig:sub2}
  \end{minipage}
   \begin{minipage}[t]{0.245\textwidth}
    \centering
    \includegraphics[width=\linewidth]{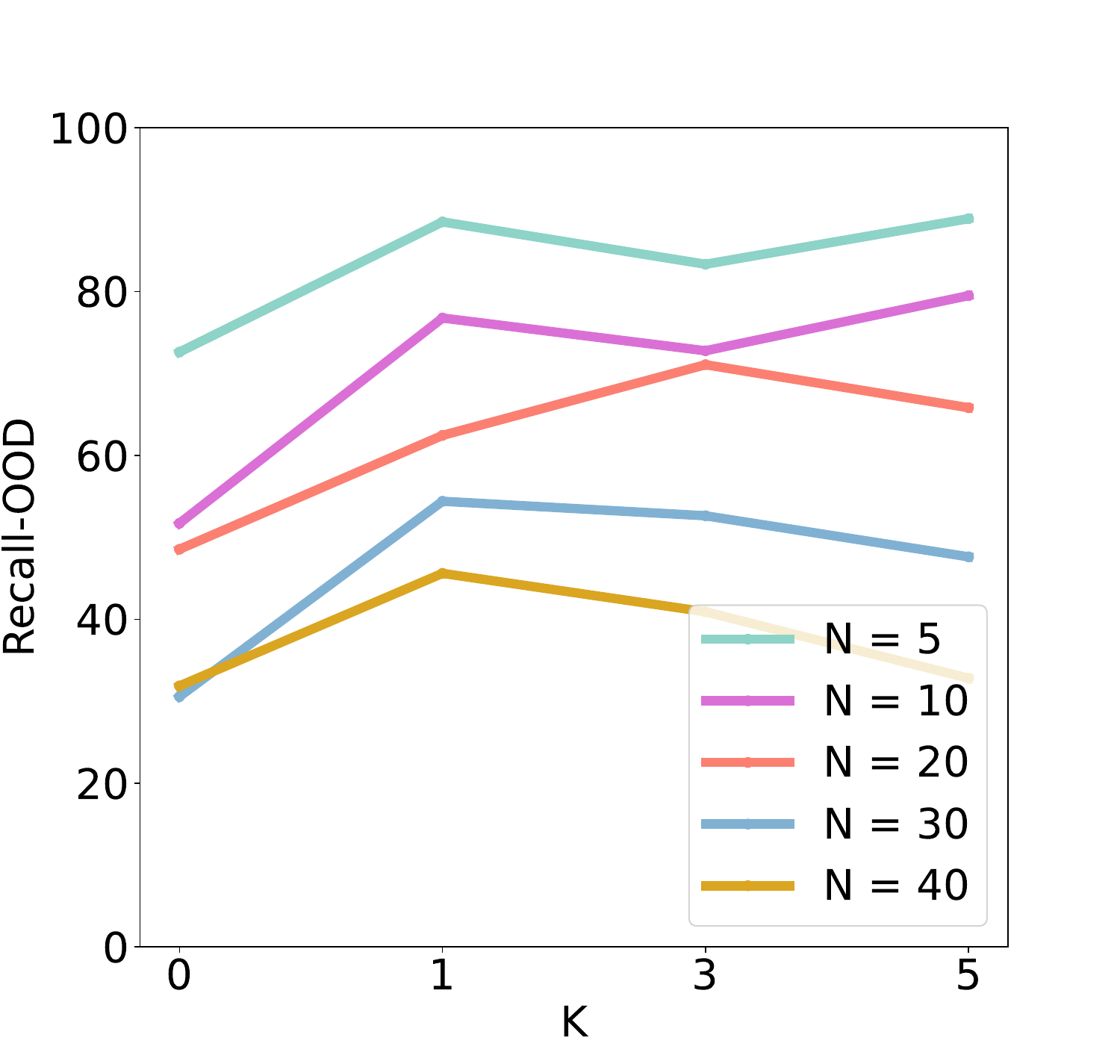}
    \subcaption[]{}
    \label{fig:sub3}
  \end{minipage}
 \begin{minipage}[t]{0.245\textwidth}
    \centering
    \includegraphics[width=\linewidth]{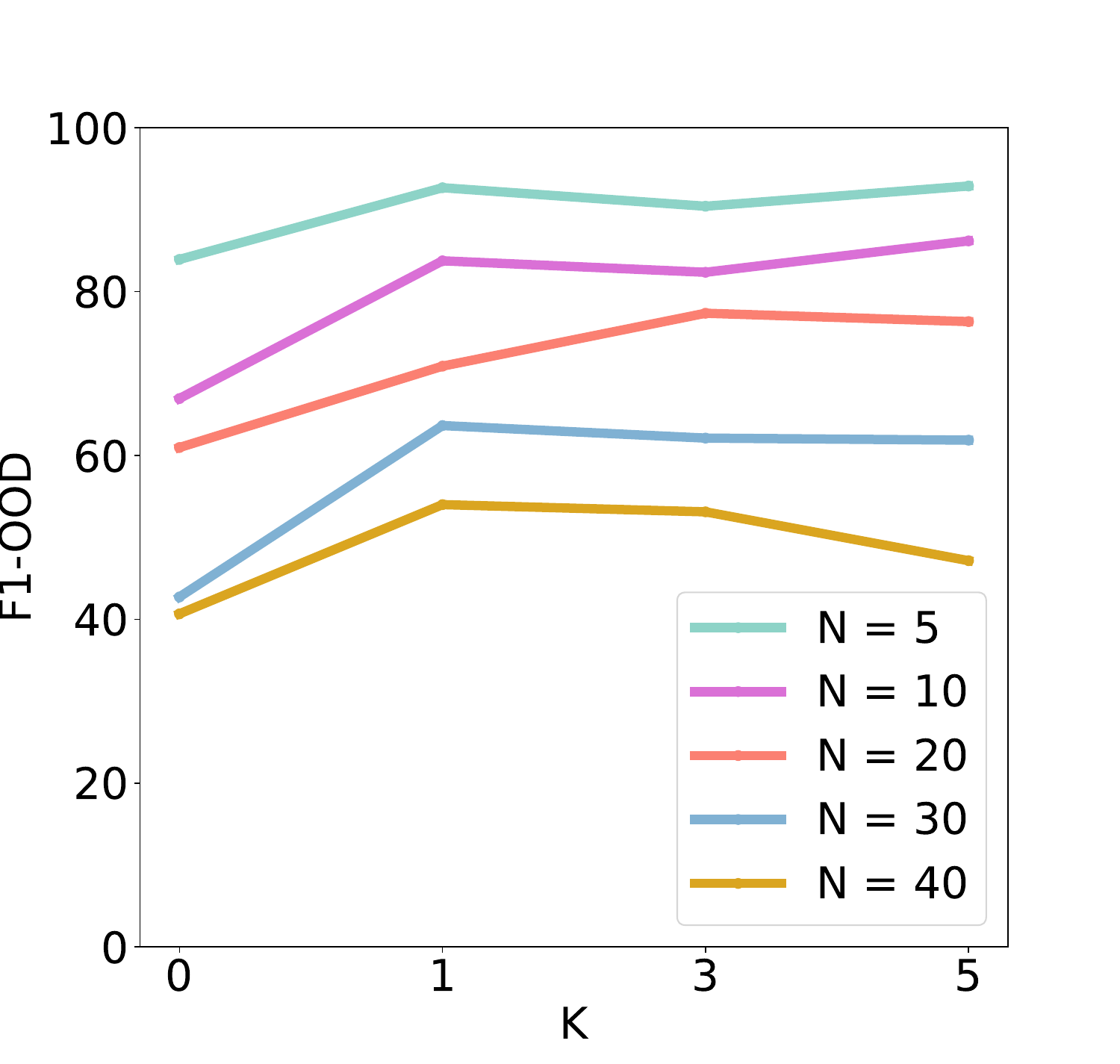}
    \subcaption[]{}
    \label{fig:sub4}
  \end{minipage}
  \caption{The effect of few-shot on different IND number. We show the changes of four metrics under different demonstration quantities. Due to the limitation of ChatGPT's input length, we conduct three sets of experiments with 1-shot, 3-shot, and 5-shot settings.}
  \label{fig:few-shot}
\end{figure*}
\subsection{FSD-LLM Results}
\label{sec:fsd_result}
Due to the length limitation of ChatGPT’s conversations, we reduce the number of IND intents and randomly select N=5,10,20,30,40 intents as IND intents, with the number of OOD intents fixed at 20. Under each setting, we test four groups of experiments with K=0,1,3,5 (K is the number of samples provided for each intent). We show the detailed FSD-LLM results in Table \ref{tab:num_few_shot} and the changing trend in Figure \ref{fig:few-shot}. We discover that:

(1) \textbf{FSD-LLM demonstrates strong competitiveness compared to the baseline in situations with a limited number of INDs} 

When N = 5, K = 5, ChatGPT outperforms UniNL by 0.76 and 3.16 on ALL-ACC and ALL-IND respectively. When N = 10 and N = 20, ChatGPT is superior to UniNL in IND classification, but inferior in UniNL in OOD detection. When N = 30 and N = 40, UniNL widens the gap with ChatGPT.

(2) \textbf{The more the number of intents, the more demonstrations are needed for IND intent recognition.}

From Figure \ref{fig:few-shot}, We find that when N=5,10, FSD-OOD achieves better F1-OOD and F1-IND performance at K=1. Even ZS-LLM achieves best ACC-IND. However, as N increases to 30 or 40, both ACC-IND and F1-IND show an upward trend with the increase of K. This suggests that the more the number of intents, the more prior knowledge about intents is needed to help distinguish between different intents. 

(3) \textbf{Too many demonstrations may introduce noise into OOD detection.}

OOD-Recall shows an overall trend of initially increasing and then decreasing. This demonstrates the model's negative transfer from IND to OOD data which means . We speculate that this could be due to significant feature distribution differences between the IND and OOD data, making it challenging for the model to learn useful features from the IND samples for OOD data.

\begin{table}
\centering
\resizebox{0.5\textwidth}{!}{%
\begin{tabular}{c|c|cc|cc|cc}
\hline
\multirow{2}{*}{\textbf{IND num}} &
  \multicolumn{1}{c|}{\multirow{2}{*}{\textbf{Few-shot}}} &
  \multicolumn{2}{c|}{\textbf{ALL}} &
  \multicolumn{2}{c|}{\textbf{IND}} &
  \multicolumn{2}{c}{\textbf{OOD}} \\ \cline{3-8} 
 &
  \multicolumn{1}{c|}{} &
  \multicolumn{1}{c}{\textbf{ACC}} &
  \multicolumn{1}{c|}{\textbf{F1}} &
  \multicolumn{1}{c}{\textbf{ACC}} &
  \multicolumn{1}{c|}{\textbf{F1}} &
  \multicolumn{1}{c}{\textbf{ACC}} &
  \multicolumn{1}{c}{\textbf{F1}} \\ \hline
    \multirow{5}{*}{\textbf{5}}  & 0 & 77.19 & 69.06 & \textbf{96.36} & 66.09 & 72.61 & 83.92 \\
                             & 1 & 88.80 & 80.11 & 90.00 & 77.60 & 88.50 & 92.67 \\
                             & 3 & 85.89 & 77.02 & 96.02 & 74.34 & 83.33 & 90.41 \\
                             & 5 & \textbf{89.11} & \textbf{81.26} & 90.01 & \textbf{78.93} & 88.89 & \textbf{92.88} \\ \cline{2-8}
                             & UniNL & 88.35	&78.1	&78.35	&75.21	&\textbf{90.85}	&92.58 \\ \hline
\multirow{5}{*}{\textbf{10}} & 0 & 63.44 & 64.85 & 86.87 & 64.64 & 51.71 & 66.94 \\
                             & 1 & 79.19 & 74.75 & 84.00 & 73.85 & 76.77 & 83.75 \\
                             & 3 & 78.81 & 76.67 & \textbf{91.00} & 76.10 & 72.77 & 82.35 \\
                             & 5 & 82.89 & \textbf{88.17} & 89.80 & \textbf{79.57} & 79.50 & 86.18 \\ \cline{2-8}
                             &UniNL &\textbf{84.17}	&78.11	&74.55	&77.1	&\textbf{89.05}	&\textbf{88.25} \\
                             \hline
\multirow{5}{*}{\textbf{20}} & 0 & 61.76 & 63.01 & 74.78 & 63.11 & 48.54 & 60.96 \\
                             & 1 & 70.89 & 72.29 & 79.29 & 72.36 & 62.44 & 70.89 \\
                             & 3 & 75.95 & 76.30 & 80.81 & 76.24 & 71.07 & 77.35 \\
                             & 5 & 77.84 & \textbf{79.57} & \textbf{88.27} & \textbf{79.74} & 65.82 & 76.33 \\ \cline{2-8}
                             &UniNL &\textbf{80.51}	&78.87	&73.21	&78.71	&\textbf{87.89}	&\textbf{81.92}\\ \hline
\multirow{5}{*}{\textbf{30}} & 0 & 56.91 & 58.13 & 74.31 & 58.64 & 30.56 & 42.72 \\
                             & 1 & 69.06 & 72.00 & 78.64 & 72.28 & 54.40 & 63.64 \\
                             & 3 & 72.18 & 72.65 & 84.31 & 73.00 & 52.63 & 62.11 \\
                             & 5 & 74.74 & 81.38 & \textbf{91.95} & 82.03 & 47.62 & 61.86 \\ \cline{2-8}
                             &UniNL & \textbf{80.76}	&\textbf{83.07}	&82.53	&\textbf{83.29}	&\textbf{78.09}	&\textbf{76.61}\\ \hline
\multirow{5}{*}{\textbf{40}} & 0 & 55.80 & 60.26 & 67.25 & 60.75 & 31.86 & 40.66 \\
                             & 1 & 66.32 & 70.66 & 76.53 & 71.08 & 45.60 & 53.99 \\
                             & 3 & 69.08 & 75.32 & 83.12 & 75.87 & 40.91 & 53.11 \\
                             & 5 & 69.35 & 75.84 & 86.84 & 76.55 & 32.80 & 47.15 \\ \cline{2-8} 
                             &UniNL & \textbf{83.69}	&\textbf{87.77}	&\textbf{86.86}	&\textbf{88.05}	&\textbf{77.32}	&\textbf{76.53}\\ \hline
\end{tabular}%
}
\caption{Performance of ChatGPT under different few-shot settings with varying five sets of IND numbers.}
\label{tab:num_few_shot}
\end{table}

\section{Qualitative Analysis}
\label{sec:qualitative}

\subsection{Effect of IND intent number}
\label{sec:ind_num}
\begin{figure}[t]
    \centering
    \begin{subfigure}[t]{0.235\textwidth}
        \includegraphics[width=\textwidth]{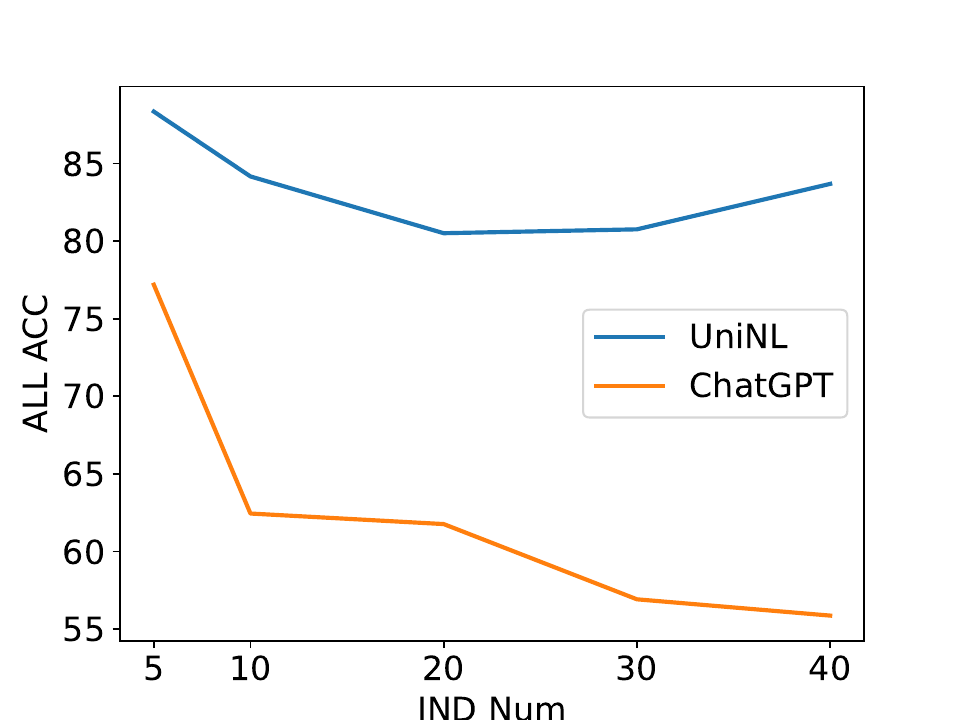}
        \caption{ALL-ACC}
        \label{fig:num_all_acc}
    \end{subfigure}
    \hfill
    \begin{subfigure}[t]{0.235\textwidth}
        \includegraphics[width=\textwidth]{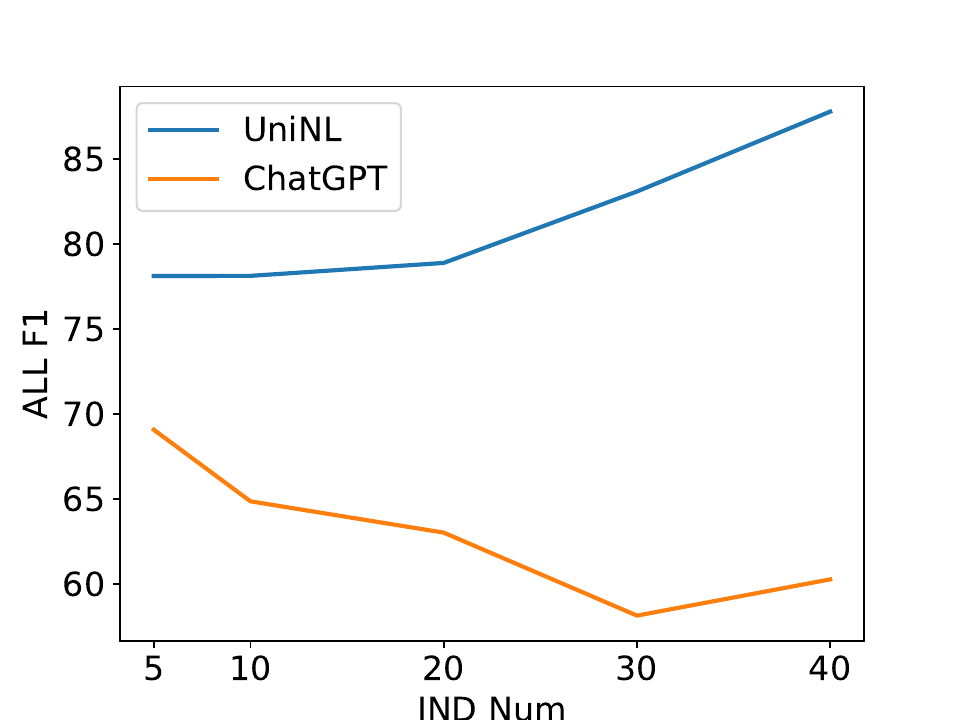}
        \caption{ALL-F1}
        \label{fig:num_all_f1}
    \end{subfigure}
        \begin{subfigure}[t]{0.235\textwidth}
        \includegraphics[width=\textwidth]{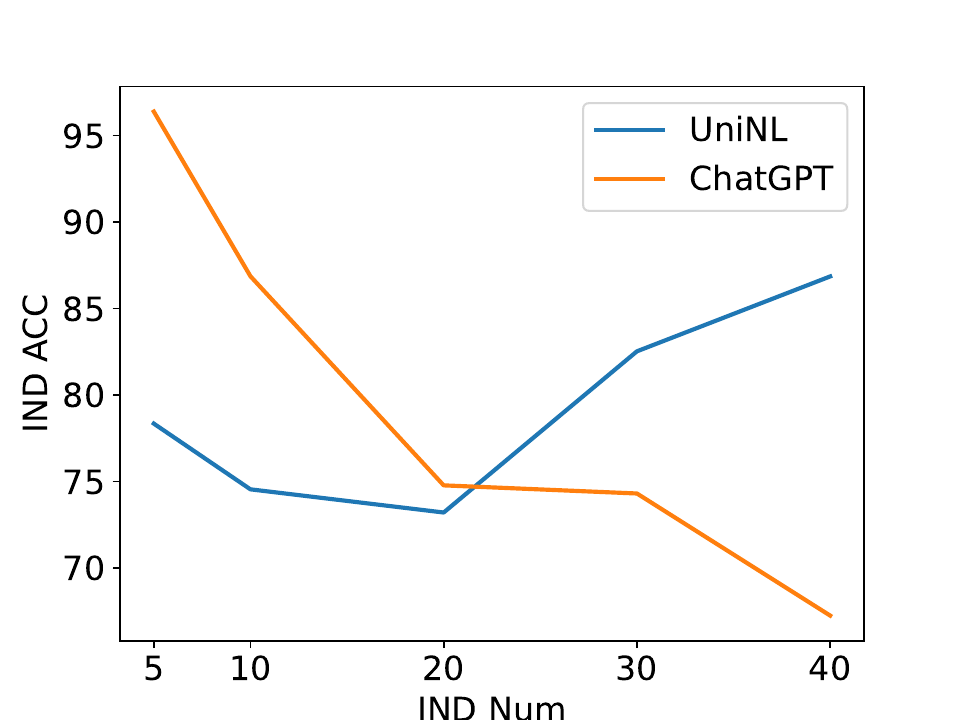}
        \caption{IND-ACC}
        \label{fig:num_ind_acc}
    \end{subfigure}
    \hfill
    \begin{subfigure}[t]{0.235\textwidth}
        \includegraphics[width=\textwidth]{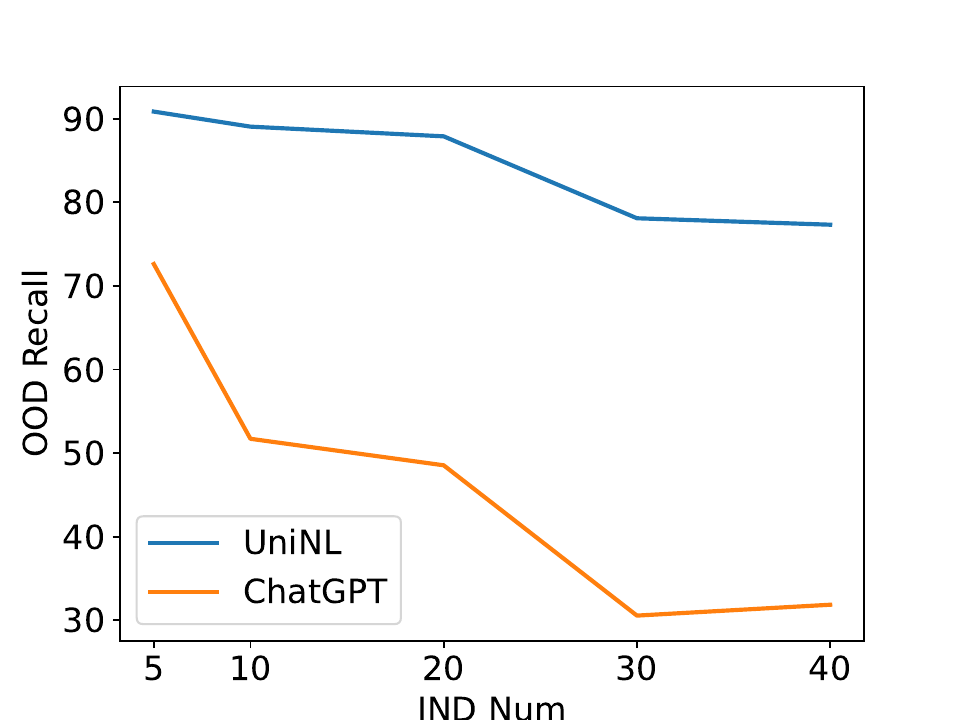}
        \caption{OOD-Recall}
        \label{fig:num_ood_recall}
    \end{subfigure}
    \caption{Changes in the ALL-ACC, ALL-F1, IND-ACC and OOD-Recall of ChatGPT and UniNL as the number of IND intent increases for banking-50\%.}
    \label{fig:ind_num}
\end{figure}

In Section \ref{sec:fsd_result}, we observe the varying performance of ChatGPT under different numbers of IND intents. In this section, we provide a detailed analysis of the changes in the effectiveness of the ZSD-LLM methods as N increases. Figure \ref{fig:ind_num} shows the trend of the changes. The results reveal that as the number of IND intents increases:  

(1) \textbf{ChatGPT is sensitive to the number of intents compared with UniNL.} As shown in Figure \ref{fig:num_all_acc} and \ref{fig:num_all_f1}, ChatGPT has the best OOD detection performance when N=5, but as the number of intents increases, both metrics consistently decrease. When it reaches 30 and 40, ACC and F1 decrease by 21.33 and 8.8, respectively. However, UniNL consistently demonstrates robust results across all numbers. UniNL still achieves an ACC of 83.69 and F1 of 87.77 when N=40.

(2) \textbf{The increase in intents leads to more severe confusion between labels.} Figure \ref{fig:num_ind_acc} shows a continuous decrease in IND-ACC, indicating that more IND samples are misclassified. We find that compared to IND samples being misclassified as OOD, the proportion of being misclassified as incorrect IND intents is increasing. This may be due to ChatGPT's understanding of intent labels not aligning with human-defined labels, leading to confusion between different label meanings. As the number of intents increases, this confusion intensifies.

(3) \textbf{The increase in intents causes a sharp drop in the OOD-recall rate.} Figure \ref{fig:num_ood_recall} shows that with the increase in the number of intents, OOD samples are more likely to be misclassified as IND. This is because the increase in IND intents number introduces more interference to ChatGPT's OOD detection. 

We believe that the advantage of ChatGPT over discriminative models lies in its OOD detection with fewer intents, where it can accurately make judgments based on its internal knowledge. However, as the number of labels increases, the confusion in label meanings becomes more prominent, resulting in a decline in both IND intent classification and OOD sample detection.

\subsection{The robustness of ChatGPT OOD detection}
\begin{figure}[t]
  \centering
  \includegraphics[width=0.5\textwidth]{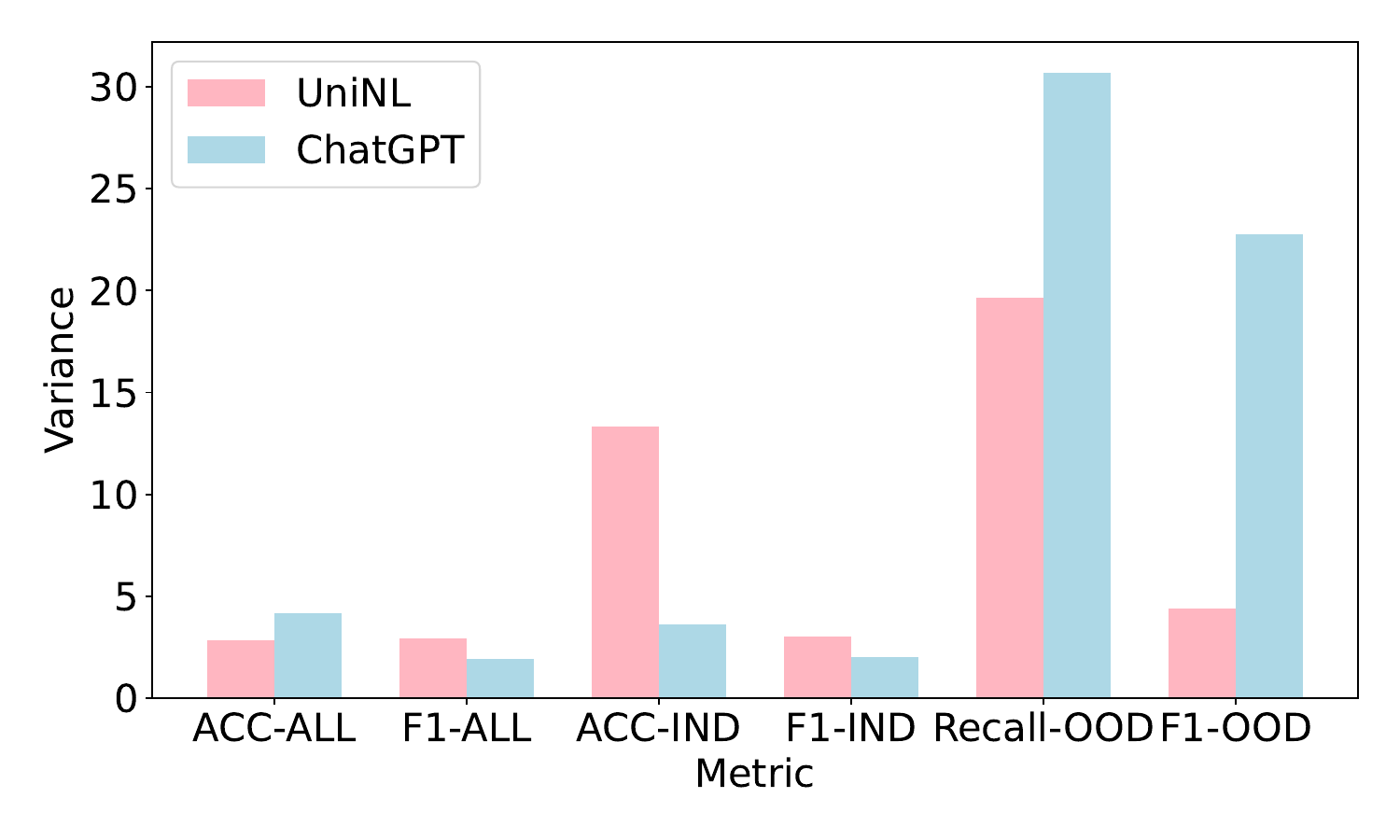}
  \caption{The variances of ChatGPT and UniNL on various metrics across five different data splits.}
  \label{variance}
\end{figure}

The robustness of OOD detection can be reflected in its ability to maintain stable across different intent partitions. To verify it, we randomly select five different IND intent set split (using five different seeds when selecting IND intents) for the experiment. Figure \ref{variance} shows the variance of the results from the five sets of experiments. We observe that UniNL and ChatGPT perform similarly. However, in terms of the IND metrics IND-ACC and IND-F1, UniNL shows greater fluctuations. In terms of OOD metrics, ChatGPT is inferior to UniNL. This highlights the differences between the two models in performing OOD intent detection. ChatGPT excels in IND intent recognition, but its stability in OOD detection is relatively poor.

\subsection{Comparison of Different LLMs}
\label{sec:llms}
\begin{table}
\centering
\resizebox{0.499\textwidth}{!}{
\begin{tabular}{l|c|c|c|c|c|c}
\hline
\multirow{2}{*}{Model} & \multicolumn{2}{c|}{ALL}               & \multicolumn{2}{c|}{IND}               & \multicolumn{2}{c}{OOD}               \\ \cline{2-7} 
                                & \multicolumn{1}{c|}{ACC} & F1 & \multicolumn{1}{c|}{ACC} & F1 & \multicolumn{1}{c|}{Recall} & F1 \\ \hline

text-davinci-002 & 54.24                        & 60.72                       & 73.6                         & 60.85                       & 34.38                           & 48.53                       \\ \hline
text-davinci-003 & 55.87                        &\underline{64.14}              & \underline{79.03}               & \underline{65.15}              & 30.81                           & 43.98                       \\ \hline
Claude           & 56.58                        & 52.76                       & 70.72                        & 52.74                       & 43.59                           & 53.12                       \\ \hline
Llama2-70b-Chat &55.5 &57.8 &67.0 &57.84 &44.0 &56.96                    \\ \hline
ChatGPT          & \underline{61.76}               & 63.01                       & 74.78                        & 63.11                       & \underline{48.54}                  & \underline{60.96}              \\ \hline
GPT4 &\textbf{68.5} & \textbf{73.55} &\textbf{87.5} &\textbf{74.07} &\textbf{49.5} &\textbf{63.26}       \\ \hline
\end{tabular}
}
\caption{OOD detection performance of six different LLMs.}
\label{tab:llms}
\end{table}
We do ZSD-LLM on other mainstream LLMs and compare them with ChatGPT.
\begin{itemize}
[itemsep=4pt,topsep=0pt,parsep=0pt,leftmargin=9pt]
\item \textbf{Text-davinci-002, text-davinci-003}\footnote{https://platform.openai.com/docs/models} belong to InstructGPT and text-davinci-003 is an improved version of text-davinci-002. Compared with GPT-3, the biggest difference of InstructGPT is that it is fine-tuned for human instructions.
\item \textbf{Claude} is an artificial intelligence chatbot developed by Anthropic\footnote{https://www.anthropic.com/product}.
\item \textbf{Llama2-70B-Chat}\footnote{https://huggingface.co/meta-llama/Llama-2-70b-chat-hf} is developed and open-sourced by Meta AI. Llama-2-70b-Chat is the native open-source version with high-precision results.
\item \textbf{GPT4}\footnote{https://platform.openai.com/docs/models} is the latest and most advanced multimodal large model from OpenAI. GPT-4 can generate more factual and accurate statements than GPT-3.5 and other language models.
\end{itemize}

\begin{figure*}[t]
  \centering
  \includegraphics[width=1\textwidth]{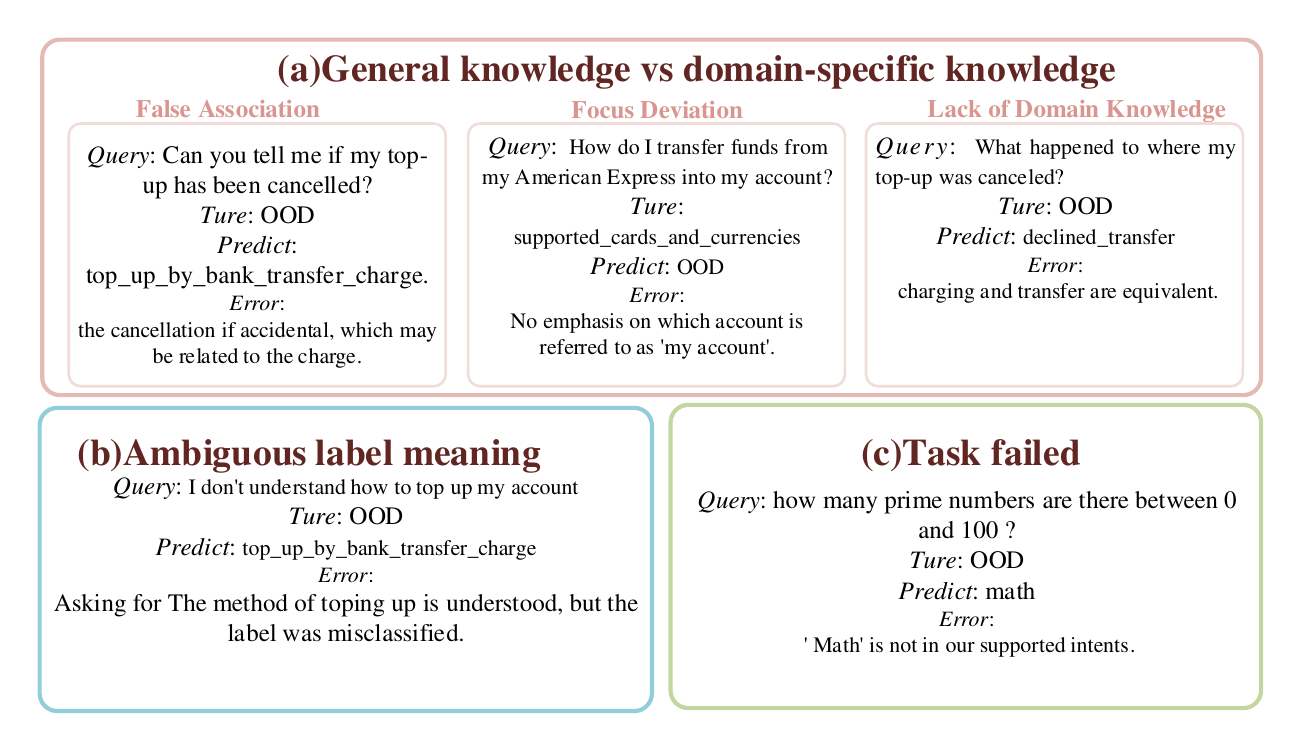}
  \caption{We identify three challenges that ChatGPT faces when performing intent OOD detection. We further subdivide the categories of \textit{general knowledge vs domain-specific knowledge} into \textit{False Association}, \textit{Focus Deviation}, and \textit{Lack of Domain Knowledge}. We provide specific cases to illustrate these errors, including the query, true label, predicted label, and the reasons for ChatGPT's misclassification.}
  \label{case}
\end{figure*}

Results are shown in Table \ref{tab:llms}. GPT4 leads in all six metrics compared to other models. Surprisingly, GPT3 performs better in IND intent recognition, with text-davinci-003 even surpassing ChatGPT and text-davinci-002 shows similar performance to ChatGPT in IND metrics. However, ChatGPT exhibits significantly better results than the GPT-3 series in OOD metrics. The differences in performance may be attributed to ChatGPT's inclusion of SFT (Supervised Fine-Tuning) during the optimization phase, which gives it an advantage in understanding human instructions. In contrast, the GPT-3 series is slightly inferior in understanding tasks, making it more inclined towards intent detection tasks. Llama2-70b-Chat and Claund also exhibit a similar phenomenon to ChatGPT, but overall, ChatGPT outperforms Llama2-70B-Chat and Claund.

\subsection{Effect of different prompts}
\label{sec:diff_prompt}
\begin{table}[t]
\resizebox{0.5\textwidth}{!}{%
\begin{tabular}{c|c|c|c|c|c|c}
\hline
\multicolumn{1}{c|}{\multirow{2}{*}{Model}} & \multicolumn{2}{c|}{ALL}                                    & \multicolumn{2}{c|}{IND}                                    & \multicolumn{2}{c}{OOD}                                    \\ \cline{2-7} 
                                              & \multicolumn{1}{c|}{ACC} & \multicolumn{1}{c|}{F1} & \multicolumn{1}{c|}{ACC} & \multicolumn{1}{c|}{F1} & \multicolumn{1}{c|}{Recall} & \multicolumn{1}{c}{F1} \\ \hline                      prompt.original &      53.10 &51.32 &56.82 &51.18 &49.47 &56.62                   \\ \hline                      prompt.detector &51.07 &53.20 &61.61 &53.25 &40.78 &51.11                         \\ \hline                     prompt.discovery     &45.63 &51.05 &63.29 &51.32 &28.39 &40.66                                \\ \hline                      prompt.order    &     42.47 & 48.57 &51.75 &48.74 &33.40 &42.09         \\ \hline prompt.reason &48.67 &47.65 &55.47 &47.56 &42.03 &50.92 \\ 
 
\hline
\end{tabular}
}
\caption{The performance of ChatGPT on various prompts.}
\label{tab:diff_prompt}
\end{table}

Prompt engineering is a crucial strategy for LLM. To verify the impact of different prompts, we devise four additional variations. They are: 

\textbf{prompt.detector}: Modify the role positioning of the LLM from intent detector to OOD detector.

\textbf{prompt.discovery}: Adopt a new task description for intent discovery. Directly make the LLM return the labels of OOD intent.

\textbf{prompt.order}: Change the order of each part, using <Utterance for test> <Task description> <Utterance for test>. 

\textbf{prompt.reason}: Output the reason before outputting the results. 

As the version of ChatGPT changes over time, we choose to use the open-source Llama2-70B-Chat for the prompt experiment. Results are shown in the Table \ref{tab:diff_prompt}. We find that in both detection and discovery modes, LLM tends to perform IND detection rather than OOD detection. The prompt.order led to a decline in performance, which may be due to the overly long instructions causing LLM to forget the information from the beginning sentence. The Reason mode exacerbate the LLM's use of incorrect domain knowledge. About the exploration of a better prompt, we leave it for future research.

\section{Challenge \& Further Disscussion}
\label{sec:discussion}
Based on the above experiments and analysis, we identify the challenging scenarios that LLMs encounter and offer guidance for future reference.

\subsection{Conflict between Domain-Specific Knowledge and General Knowledge}\

The majority of errors are caused by the model's incorrect utilization of certain knowledge, which may be due to discrepancies between LLM and humans in both intent and task understanding. We refer to it as conflicts between generic knowledge within the model and domain-specific knowledge required for the task. Specifically, they result in three types of errors: \textbf{false association}, \textbf{focus deviation}, and \textbf{lack of domain knowledge}.We display the relevant cases in Figure \ref{case}(a). 

Our FSD-LLM can inject domain-specific knowledge in Section \ref{sec:fsd_result}. However, the effectiveness of demonstrations seems to be influenced by various factors such as the numbers of IND intents, the number of demonstrations. Besides, the knowledge conflict may be more pronounced in larger models than in smaller ones (Section \ref{sec:llms}), so how to inject domain-specific knowledge into LLM and eliminate noise interference from useless general knowledge will be a future research direction. 

\subsection{Difficulty of Knowledge Transfer from IND to OOD}

Experiments in Section \ref{sec:fsd_result} show that FSD-LLM achieve limited improvements in OOD sample detection. Ambiguous label meaning is quite common as shown in Figure \ref{case}(b). Too many demonstrations even lead to a decrease in results especially when the number of intents is low. One improvement method is to add OOD samples in prompt, but estimating the number of OOD intents and providing sufficiently comprehensive OOD data is challenging. Future research can focus on \textbf{how to enable models to learn transfer knowledge from IND's prior knowledge to OOD detection.} One optimization direction is to focus on how to select high-quality examples to inject diverse and noise-free prior knowledge.

\subsection{Sensitivity to input length}
When it comes to OOD intent detection  with a large number of intents, ChatGPT exhibits errors in understanding instructions as mentioned in Section \ref{sec:main_result}. When the prompt becomes longer, it is easy to exceed the processing capacity of LLMs, leading to erroneous outputs as shown in Figure \ref{case}(c). This limitation restricts the versatility of models like ChatGPT in handling diverse task scenarios and calls for future research efforts to address this issue.
\subsection{Future Insights}
We look forward to the emergence of LLM that can set new benchmarks and open up new areas of application. Unfortunately, such models have not yet appeared. On the contrary, we discovery some shortcomings in their ability to handle OOD tasks. At the same time, we also find the powerful zero-shot and few-shot capabilities of LLMs. Through the analysis, we believe that the further improvement directions for LLMs are \textbf{1) injecting domain knowledge, 2) strengthening knowledge transfer from In-Distribution (IND) to OOD, and 3) understanding long instructions}. We hope that our work can bring a deeper understanding of LLMs to the academic community, and we look forward to future work that can improve the application of LLMs in domain tasks.

\section{Conculsion}
In this paper, we conduct a comprehensive evaluation of ChatGPT for OOD intent detection. We first compare the performance of ChatGPT with traditional discriminative models and identify a significant performance gap. Additionally, we observe that ChatGPT excels in handling tasks with a small number of intents but struggles with tasks involving a large number of intents. While incorporating demonstration examples shows some improvements, there is still considerable room for enhancement. We recommend future research to focus on improving large-scale models for OOD tasks by incorporating domain-specific knowledge into the models and how to learn transfer relationship from OOD detection.
\section{Limitations}
In this paper, we investigate the advantages, disadvantages and challenges of LLMs in open-domain intent OOD detection. Although we conduct extensive experiments, there are still several directions to be improved: (1) We propose FSD-LLM to do few-shot OOD detection for LLM, but the demonstration examples are randomly selected. In this paper, we do not consider the diversity and quality of the demonstration examples. (2)We use closed source LLMs in this paper like ChatGPT and GPT4. Although we ensure that all experiments are based on the same version(gpt-3.5-turbo-0301, gpt-4-0613), further updates of ChatGPT may lead to results in the future that will differ from those reported in this paper. 

\section{Acknowledgements}
Thanks to the State Key Laboratory of Massive Personalized Customization System Technology for their support. This work was supported by the National Natural Science Foundation of China (NSFC No.62076031 and No.62076036).


\nocite{*}
\section{Bibliographical References}\label{sec:reference}

\bibliographystyle{lrec-coling2024-natbib}
\bibliography{lrec-coling2024-example}


\end{document}